# Hierarchy Influenced Differential Evolution: A Motor Operation Inspired Approach

Shubham Dokania[1], Ayush Chopra[1], Feroz Ahmad[1] and Anil Singh Parihar[1]

[1]*Delhi Technological University, New Delhi, India*
shubham.k.dokania@gmail.com, {ayushchopra_2k14, ferozahmad_2k14, anil}@dtu.ac.in

Keywords: Differential Evolution, Metaheuristics, Continuous Optimization, Hierarchical Influence

Abstract: Operational maturity of biological control systems have fuelled the inspiration for a large number of mathematical and logical models for control, automation and optimisation. The human brain represents the most sophisticated control architecture known to us and is a central motivation for several research attempts across various domains. In the present work, we introduce an algorithm for mathematical optimisation that derives its intuition from the hierarchical and distributed operations of the human motor system. The system comprises global leaders, local leaders and an effector population that adapt dynamically to attain global optimisation via a feedback mechanism coupled with the structural hierarchy. The hierarchical system operation is distributed into local control for movement and global controllers that facilitate gross motion and decision making. We present our algorithm as a variant of the classical Differential Evolution algorithm, introducing a hierarchical crossover operation. The discussed approach is tested exhaustively on standard test functions as well as the CEC 2017 benchmark. Our algorithm significantly outperforms various standard algorithms as well as their popular variants as discussed in the results.

## 1 Introduction

Evolutionary algorithms are classified as meta-heuristic search algorithms, where possible solution elements span the n-dimensional search space to find the global optimum solution. Over the years, natural phenomena and biological processes have laid the foundation for several algorithms for control and optimization that have highlighted their applicability in solving intricate optimization problems. For instance, at the cellular level in the E.Coli Bacterium, there is sensing and locomotion involved in seeking nourishment and avoiding harmful chemicals. These behavioral characteristics fuelled the inspiration for the Bacterial Foraging Optimization algorithm (Passino, 2002)(Onwubolu and Babu, 2013). Particle Swarm Optimization (Kennedy and Eberhart, 1995) is a swarm intelligence algorithm based on behavior of birds and fishes that models these particles as they traverse an n-dimensional search space and share information in order to obtain global optimum. From a biological control point, the human brain represents one of the most advanced architectures and several research attempts seek to mimic its functional accuracy, precision and efficiency. The brain function activities can be broadly classified into 2 categories: sensory and motor operations. Sensory cortical functions inspired the concept of neural networks that are being scaled successfully in deep learning to solve vast amount of problems.

The human motor function represents a distributed neural and hierarchical control system. It can be clas-sified as having local control functions for movement as well as higher level controllers for gross motion and decision making. The execution of motor operation involves distributed brain structures at different levels of hierarchy. These include the pre-frontal cortex, motor cortex, spinal cord, anterior horn cells etc (Shaw et al., 1982). For executing an action sequence, a sequence of actions is implemented by a string of subsequences of actions each implemented in a different part of the body. The operational structure has been depicted in Figure 1(Passino, 2005). For optimality of actions, neurons act in unison. The neurons in the motor cortex act like global leaders and send inhibitory or facilitatory influence over anterior horn cells, the local leaders, located in the spinal cord(Shaw et al., 1982). These local leaders are connected to muscle fibers, the effectors, through a peripheral nerve and neuromuscular junction. Efficient execution of task requires feedback based facilitation and inhibition of the effectors over the anterior horn cells. These sequence of operations realise the optimal convergence of the system leading to smooth motor execution.

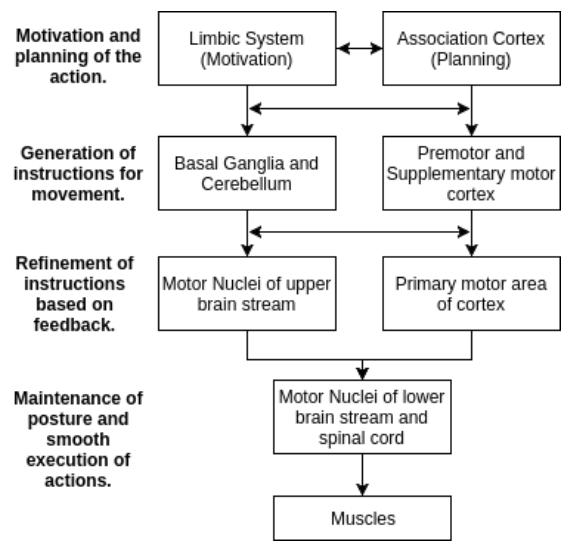

Figure 1: Hierarchy of Motor Control in Humans

The present work introduces an algorithm modelled intuitively on the distributed and hierarchical operation of the brain motor function.

The Classical DE Algorithm (Storn and Price, 1995), proposed by Storn and Price has been hailed as one of the premier evolutionary algorithms, owing to its simple yet effective structure(Das and Suganthan, 2011). However, in recent times, it has been criticized for its slow convergence rate and inability to effectively optimize multimodal composite functions(Das and Suganthan, 2011). This work focusses on supplementing the algorithm's performance through the introduction of hierarchical influence in the pipeline. The architecture enables the algorithm to control the flow of agents through the cummulative effect of global and local leaders in the hierarchy.
The proposed approach, Hierarchy Influenced Differential Evolution (HIDE), has been subjected to exhaustive analysis on the hybrid and composite objective functions of the CEC 2017 benchmark(Awad et al., 2016). Comparison with the classical DE algorithm and its other popular variants including JADE and PSODE (Zhang and Sanderson, 2009) highlights the particular viability of the schemed approach in solving complex optimization tasks. We show that even with fixed parameters, HIDE is able to outperform adaptive architectures such as JADE by a respectable margin, as discussed in the result sections.

## 2 Classical Differential Evolution

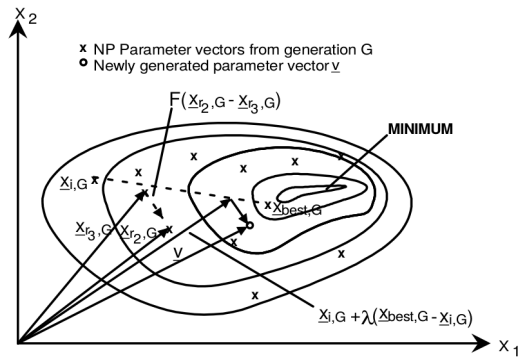

Figure 2: Motion planning of individuals in DE on two dimensional example of objective function.

The classical Differential Evolution (DE) algorithm is a population-based global optimization algorithm, utilizing a crossover and mutation approach to generate new individuals in the population for achieving optimum solutions(Das and Suganthan, 2011). For each individual $x_i$ that belongs to the population for generation $G$, DE randomly samples three individuals from the population namely $x_{r1,G}$, $x_{r2,G}$ and $x_{r3,G}$. Employing these randomly chosen points, a new individual trial vector, $v_i$, is generated using equation (1):

$$v_i = x_{r1,G} + F(x_{r2,G} - x_{r3,G}) \qquad (1)$$

Where, $F$ is called the differential weight (Usually lies between $[0,1]$).
To obtain the updated position of the individual, a crossover operation is implemented between $x_{i,G}$ and $v_i$, controlled by the parameter $CR$ called the crossover probability. The value for $CR$ always lies between $[0,1]$.

## 3 Hierarchy Influenced Differential Evolution

Taking inspiration from the human motor system, we model the hierarchical motor operations in our optimization agents, where we define a global leader which influences the action of several distributed local leaders and the particle agents which act as the effectors. The global leader is analogous to the decision making and planning section in the motor system hierarchy whilst, the local leaders correspond to motion generators acting under the influence of the global leader.
The position of each particle in the population is affected by the influence of global leader and lo-cal leaders, while also being affected by a randomly chosen particle from the population to induce some stochasticity in the optimization pipeline. We first model the influence of the global leader on the local leaders and the influences of the local leaders on each population element using equation (3) and (4). We introduce a hierarchical crossover between the two influencing equations governed by a hierarchical crossover parameter *HC*.

Analogously to the brain motor operation as depicted in Figure 1, the update of particle positions requires generating feedback for the leaders as a part of the optimization procedure, and hence the local leaders and the global leader are updated based on their objective function value generated from the perturbations in population particles. This series of events comprise of one optimization pass (one generation step). On execution of several optimization passes as described, the system is able to converge to an optimal configuration, analogous to the successful execution of the required task as shown in the final steps of Figure 1.

For each particle $x_{i,G}$, $i = 0, 1, 2, ...NP-1$ for generation G, the trial vector $x'_i$ of the particle, is governed by the hierarchical crossover operation and a mutation operation as follows :

$$u_i = \begin{cases} E_g, & if\ G < HC * G_t \\ E_l, & otherwise \end{cases} \quad (2)$$

$$E_g = g_L + F(x_{L_i,G} - x_{r,G}) \quad (3)$$

$$E_l = x_{L_i,G} + F(x_{i,G} - x_{r,G}) \quad (4)$$

for each dimension $j$ of $x_{j,i,G}$:

$$x'_{j,i} = \begin{cases} x_{j,i,G} & if\ rand(0,1) < HC \\ u_{j,i} & otherwise \end{cases} \quad (5)$$

$$x_{i,G+1} = \begin{cases} x'_{i,G}, & if\ f(x'_{i,G}) < f(x_{i,G}) \\ x_{i,G}, & otherwise \end{cases} \quad (6)$$

where,
$G_t$ is the total number of generations,
$x_{i,G+1}$ is the vector position of $x_{i,G}$ for next generation
$F$ is factor responsible for amplification of differential variation,
$f$ is the objective function,
$x_{i,G}$ is the current position of the individual for generation G,
$u_i$ is the intermediate trial vector of the current individual,
$E_g$ represents the global and local leader interaction,
$E_l$ represents the local leader and effector interaction,
$g_L$ is the global leader for generation G,
$x_{L_i,G}$ is the position of the local leader for current

---

**Algorithm 1** Hierarchy Influenced Differential Evolution

1: **procedure** START
2:   Initialize parameters (*HC*, *F*, *P*, $N_l$, *NP*).
3:   Generate initial global leader $g_L$ as a random point.
4:   Generate $N_l$ local leader points around $g_L$ global leader.
5:   Generate *NP* points for population *P* around the local leaders using a Normal distribution with identity covariance.
6:   **while** Termination criteria is not met **do**
7:     **for** each individual $x_{i,G}$ in *P* **do**
8:       Determine the corresponding local leader $x_{L_i,G}$ from the set of all local leader based on nearest position.
9:       Let $u = 0$ be an empty vector.
10:      Let *G* and $G_t$ be the current generation and total generations of the procedure.
11:      **if** $G < (HC * G_t)$ **then**
12:        $u_i = E_g$ from (3).
13:      **else**
14:        $u_i = E_l$ from (4).
15:      **end if**
16:      $x'_i$ = BinomialCrossover($u_i$, $x_{i,G}$, *CR*)
17:      **if** $f(x'_i) < f(x_{i,G})$ **then**
18:        Replace $x_{i,G}$ with $x'_i$ in the next generation.
19:      **end if**
20:    **end for**
21:    Alter local leaders in each population cluster based on objective function value.
22:    Compute updated global leader $g_L$.
23:  **end while**
24: **end procedure**

---

**Algorithm 2** Binomial_Crossover(*u*, *x*, *CR*)

1: **procedure** START
2:   Let $x' = 0$ be an empty vector.
3:   Select a random integer k = $irand(\{1,2,...,d\})$; where d = number of dimensions
4:   **for** each dimension *j* **do**
5:     **if** $random(0,1) < CR$ **or** $j == k$ **then**
6:       Set $x'_j = u_j$
7:     **else**
8:       Set $x'_j = x_j$
9:     **end if**
10:  **end for**
11: **end procedure**

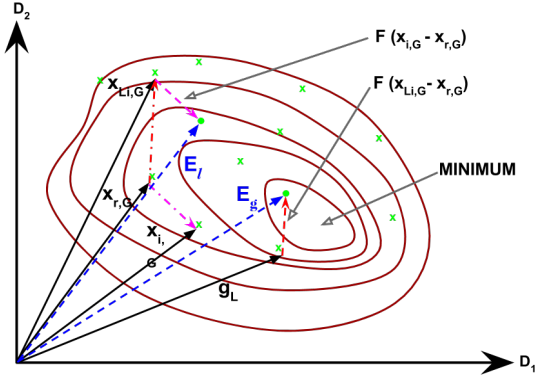

Figure 3: Hierarchical Decisive Motion planning of individuals in HIDE on two dimensional example of objective function. The position vectors resulting from the influence of global leader and local leaders are both represented as $E_g$ and $E_l$ on the contour of a two dimensional objective function.

individual,
$x_{r,G} \varepsilon$ P ; r $\varepsilon$ [0,1,.. NP-1]
$x'_{j,i}$ is the trial vector

$x_{r,G}$ is randomly chosen particle from the population to induce stochasticity. The hierarchical operation is affected by the global leader $g_L$ and the local leader $x_{L_i,G}$ through the parametric equations (3) and (4). Switching between the two is governed by the hierarchical crossover parameter *HC*.

## 3.1 Hierarchical Crossover

Convergence trend in HIDE is largely pivoted about (3) and (4), which in unison, lend a hierarchical structure to the algorithm. A successful optimization algorithm involves establishing a trade-off between exploration and exploitation. Achieving global optimization can be visualized as collaboration of two forces, exploration over a larger subspace followed by intensive exploitation over the resulting search space governed by clusters. Phase 1, involving (3) is marked by the interaction between the global and local leaders representing decision planning and facilitation of gross motion. This is followed by phase 2, involving (4) wherein the local leaders interact with and guide their effector population to control intricate motion over the constraint subspace to achieve smooth convergence. Robust covergence necessitates an optimal transition from phase 1 to phase 2 in the hierarchy. This hierarchical transition is characterized by our proposed parameter, HC. The value of HC belongs to [0,1]. An optimal value for HC was observed experimentally to lie about one-quarter. For the purpose of our experiments, we have fixed *HC* to be 0.27. Thus, this defines a deterministic cut after 27% of the total generation budget. The crossover probability defined here was observed to be mostly 50% smaller in comparison to other DE variants.

The HIDE algorithm achieves a performance improvement in the early optimization phase ($G$ ¡ $HC * G_t$) by replacing clusters of the initially generated candidate solutions with the locally best. This strategy rules out a number of mutation vectors that are more unfavorable in terms of performance gain. Additionally, by focusing on mutants of the globally best candidate solution the search space is explored rather quickly during this phase. After the population advances to $HC * G_t$ generations, the algorithm changes its reference point (the trial vector) to the locally best candidate solutions of a certain cluster. That is, having approached a closer distance from the optimal, the algorithm is able to exploit the search space. Our proposition is complemented by the observations in our results section wherein we significantly outperform several popular algorithms on involved multimodal hybrid and composite functions in higher dimensions.

## 4 Results and Discussions

All evaluations were performed using Python 2.7.12 with Scipy(Oliphant, 2007) and Numpy(Van Der Walt et al., 2011) for numerical computations and Matplotlib (Hunter, 2007) package for graphical representation of the result data. This section is divided into two sub-sections: Section A provides description

Table 1: Algorithm Parameter Settings used for comparision

| Algorithm | Parameter | Value |
|---|---|---|
| DE | $F$ | 0.5 |
|  | $CR$ | 0.9 |
| PSODE | $w$ | 0.7298 |
|  | $\phi_p$ | 1.49618 |
|  | $\phi_g$ | 1.49618 |
|  | $F$ | r $\varepsilon$ [0.9,1.0] |
|  | $CR$ | r $\varepsilon$ [0.95,1.0] |
| JADE | $p$ | 0.05 |
|  | $c$ | 0.1 |
| HIDE | $HC$ | 0.27 |
|  | $F$ | 0.48 |
|  | $CR$ | 0.9 |
|  | $N_l$ | 5 |

Table 2: Objective Function Value for Dimension: 30

| $f_{id}$ | DE best | DE mean | JADE best | JADE mean | PSO-DE best | PSO-DE mean | HIDE best | HIDE mean |
|---|---|---|---|---|---|---|---|---|
| $f_1$ | 100.001508 | 4334.43848 | 100.001338 | 100.056201 | 364.295574 | 4236.36321 | **100.0** | **100.0** |
| $f_2$ | 40412441.0 | 5.1296e+19 | **200.0** | 1535352368 | 332899.0 | 9.59068e+11 | **200.0** | **159855.5** |
| $f_3$ | 17926.8728 | 22131.5427 | 69304.9261 | 74080.7004 | 15792.5475 | 21683.2090 | **3679.81159** | **8999.94726** |
| $f_4$ | 481.255055 | 519.422652 | 403.633939 | **442.206911** | 468.341175 | 479.341966 | **400.004163** | 443.016156 |
| $f_5$ | 689.041352 | 737.79326 | **667.50756** | **735.204027** | 715.904429 | 746.548906 | 685.40454 | 738.842184 |
| $f_6$ | **643.626307** | 652.582714 | 651.39169 | 655.142819 | 642.724237 | 655.106996 | 644.701241 | **652.002395** |
| $f_7$ | 883.347367 | 962.591129 | **779.907693** | **818.344111** | 790.014281 | 854.285524 | 812.923573 | 856.90477 |
| $f_8$ | 923.37426 | 967.251501 | 931.500175 | **957.362003** | **915.414882** | 960.486239 | 930.288539 | 964.11663 |
| $f_9$ | 5652.48396 | 7878.78144 | 4953.05469 | 5146.60095 | 6018.41719 | 9042.41018 | **4003.11807** | **4734.98436** |
| $f_{10}$ | **3596.63104** | 4536.98976 | 4012.72329 | **4204.18969** | 3934.60671 | 4863.74111 | 3793.78177 | 4346.74134 |
| $f_{11}$ | 1162.40596 | 1184.63401 | 1152.74853 | 1174.58813 | 1165.14499 | 1189.17178 | **1149.74849** | **1171.13041** |
| $f_{12}$ | 56679.4351 | 317650.613 | 24821.1717 | 58930.0902 | 10221.0774 | 161046.055 | **9208.28924** | **41947.2226** |
| $f_{13}$ | 3002.02949 | 18794.8359 | 4276.90774 | 13775.8162 | 3871.27983 | 10612.2635 | **1664.06241** | **2453.60697** |
| $f_{14}$ | 1773.18079 | 5502.16038 | 1496.21986 | 42868.9158 | 1555.45276 | 4029.80853 | **1462.92685** | **1504.19151** |
| $f_{15}$ | 1860.43566 | 2484.68996 | 1688.05046 | 2222.67432 | 1651.74747 | 2223.06054 | **1611.07440** | **1852.66177** |
| $f_{16}$ | 2517.43962 | 2827.00496 | 2344.19818 | 2621.61868 | **2239.24272** | **2664.11466** | 2298.04196 | 2691.67481 |
| $f_{17}$ | 2321.17594 | 2604.52977 | 2062.89802 | 2546.99559 | 2107.43677 | 2457.34021 | **1820.80664** | **2418.72383** |
| $f_{18}$ | 38987.2824 | 94156.3285 | **11841.6081** | 184888.162 | 62294.8532 | 118430.289 | 12578.0037 | **23024.1119** |
| $f_{19}$ | 2043.46988 | 3010.23537 | 1959.71819 | 2156.95787 | 3049.52231 | 6840.40839 | **1949.27171** | **1987.86676** |
| $f_{20}$ | 2625.53915 | 2864.83261 | **2706.31444** | **2805.60006** | 2619.99649 | 2895.10724 | 2753.80621 | 2966.03579 |
| $f_{21}$ | 2412.08175 | 2504.77777 | 2414.52134 | 2456.71898 | 2431.74029 | 2478.84135 | **2200.0** | **2442.73431** |
| $f_{22}$ | 2300.48179 | 5655.56932 | **2300.0** | **4157.69878** | 2307.72135 | 6811.06916 | 2300.00998 | 6795.24842 |
| $f_{23}$ | 3050.65450 | 3572.96506 | **2772.00202** | **2946.74932** | 2764.92246 | 3199.87436 | 2883.27689 | 3543.83934 |
| $f_{24}$ | 3104.62369 | 3290.69875 | 2891.55764 | 2965.22556 | **2911.63347** | 2983.77293 | **2500.0** | 2940.75997 |
| $f_{25}$ | 2916.18065 | 2946.71175 | 2875.10684 | 2881.09138 | 2875.49884 | 2889.94367 | **2874.17111** | **2877.48490** |
| $f_{26}$ | 4043.69140 | 6756.3724 | 2900.0 | **3266.51098** | **2800.00780** | 3273.12876 | 2900.0 | 3298.49053 |
| $f_{27}$ | 3200.00585 | 3998.87649 | 3145.81035 | **3189.82261** | 3145.42523 | 3639.63413 | **3132.81628** | 3284.28897 |
| $f_{28}$ | 3290.74402 | 3326.26398 | 3195.48683 | 3131.02731 | 3195.48683 | 3225.59405 | **3100.0** | **3115.50582** |
| $f_{29}$ | 3720.31459 | 4115.18580 | **3305.31013** | **3626.88755** | 3535.95229 | 3867.59306 | 3352.84505 | 3709.10237 |
| $f_{30}$ | 3359.03076 | 3900.82666 | **3263.49653** | 3749.61072 | 3312.63502 | 3524.71447 | 3298.70464 | **3421.71532** |
| $w/t/l$ | 2/0/28 | 0/0/30 | 8/2/20 | 11/0/19 | 4/0/26 | 1/0/29 | 15/2/13 | 17/0/13 |

Table 3: Objective Function Value for Dimension: 50

| $f_{id}$ | DE best | DE mean | JADE best | JADE mean | PSO-DE best | PSO-DE mean | HIDE best | HIDE mean |
|---|---|---|---|---|---|---|---|---|
| $f_1$ | 5884574.87 | 367294248.5 | 136.072384 | 3708.75086 | 5811.21899 | 154233.646 | **106.072862** | **3665.41927** |
| $f_2$ | 4.7181e+24 | 3.3649e+44 | **2635725.0** | **5.0237e+26** | 2.2121e+19 | 2.5445e+23 | 2.2799e+17 | 1.0072e+31 |
| $f_3$ | 45520.9663 | 62237.2968 | 143481.793 | 156166.762 | 52308.4274 | 64435.2406 | **44613.2999** | **58182.8373** |
| $f_4$ | 574.400328 | 801.384952 | 418.580378 | 470.113207 | 477.080964 | 574.528479 | **400.005049** | **447.775413** |
| $f_5$ | 816.394775 | 843.258843 | 809.89948 | 834.13126 | **778.59312** | 831.066954 | 791.405194 | **830.218472** |
| $f_6$ | 652.54191 | 655.794152 | **633.21788** | **654.893828** | 653.291336 | 658.183613 | 645.25633 | 656.060597 |
| $f_7$ | 1109.02123 | 1263.03848 | **889.036574** | **944.90319** | 915.153525 | 1047.43879 | 989.957862 | 1186.2487 |
| $f_8$ | 1139.27892 | 1175.8931 | 1118.3391 | **1144.60474** | **1092.62639** | 1159.03235 | 1100.4760 | 1168.5299 |
| $f_9$ | 22196.3878 | 29218.7759 | 11958.2800 | **13174.6623** | 24753.0405 | 32233.9545 | **10251.4763** | 14752.7168 |
| $f_{10}$ | 6228.49289 | 7289.18367 | 6054.70769 | 6833.30631 | 6207.79530 | 7055.59523 | **6050.43437** | **6609.80456** |
| $f_{11}$ | 1170.85860 | 1258.51763 | 1202.69485 | 1232.20426 | 1206.15456 | 1252.93954 | **1156.4396** | **1205.2544** |
| $f_{12}$ | 677263.079 | 16987989.9 | 74784.6159 | 530814.648 | 584300.698 | 3448448.79 | 126908.215 | **494471.075** |
| $f_{13}$ | 6005.53530 | 16893.94992 | 2041.48812 | 4332.5945 | 1572.25297 | **4301.82960** | **1484.76179** | 7760.05613 |
| $f_{14}$ | 38490.5323 | 174367.450 | **2466.04705** | 238838.470 | 16327.4231 | 67939.0002 | 2967.8184 | **26290.3161** |
| $f_{15}$ | 2278.14122 | 26989.2555 | 13553.0418 | 25636.7696 | 3443.58734 | **9167.26709** | **1938.20040** | 14976.7218 |
| $f_{16}$ | 2722.02601 | 3176.91690 | **2345.40070** | **2916.56101** | 2521.89381 | 3146.04527 | 2436.44933 | 2978.37746 |
| $f_{17}$ | 2799.94977 | 3289.61565 | 2568.38357 | 2907.86927 | 2887.28110 | 3236.95792 | **2561.37030** | **2874.96503** |
| $f_{18}$ | 264037.125 | 872072.477 | 36176.5867 | **113941.317** | **26965.2851** | 114846.121 | 260540.781 | 536454.326 |
| $f_{19}$ | 10051.9124 | 20380.2571 | 2089.17225 | 7763.17234 | 9905.85082 | 16555.7569 | **2013.12690** | **3609.25896** |
| $f_{20}$ | 2950.92319 | 3274.33401 | 3041.81309 | 3113.28946 | 2991.58929 | 3361.82394 | **2495.03177** | **3080.13747** |
| $f_{21}$ | 2596.7256 | 2689.68836 | 2526.19089 | 2597.6771 | 2555.8788 | 2642.38159 | **2447.75827** | **2570.91101** |
| $f_{22}$ | 9713.99324 | 10803.6537 | 10759.5967 | 11032.8809 | 8918.43626 | 10465.0224 | **8181.4460** | **9755.0703** |
| $f_{23}$ | 3451.10494 | 4200.17442 | 2971.16064 | 3237.77866 | 2977.55496 | 3490.63975 | **2851.65025** | **3162.31362** |
| $f_{24}$ | 3434.46502 | 3682.84670 | 3103.95517 | 3185.38267 | **3036.79960** | **3158.33050s** | 3136.92774 | 3284.65609 |
| $f_{25}$ | 3141.14488 | 3292.30344 | 2931.16295 | 2962.47175 | 2931.92695 | 3008.89535 | **2931.14231** | **2954.76783** |
| $f_{26}$ | 4906.13284 | 7989.49096 | **2900.0** | 3346.87403 | 2900.44189 | 3653.75774 | **2900.0** | **3262.66849** |
| $f_{27}$ | 3200.01070 | 3792.64558 | 3143.03805 | 3184.64635 | 3158.17823 | 3397.13032 | **3141.01087** | **3176.01152** |
| $f_{28}$ | 3300.01082 | 3431.57091 | **3240.72586** | **3288.25303** | 3300.25760 | 3300.25760 | 3243.63199 | 3294.37323 |
| $f_{29}$ | 3812.47551 | 4605.34953 | **3533.94574** | **3956.83524** | 3955.32453 | 4364.18129 | 3653.67555 | 3966.47195 |
| $f_{30}$ | 3673.71196 | 5813.17375 | 3916.72571 | 4869.08933 | 3730.30935 | 5143.07870 | **3346.48367** | **4747.88675** |
| $w/t/l$ | 0/0/30 | 0/0/30 | 8/1/21 | 9/0/21 | 4/0/26 | 3/0/27 | 17/1/12 | 18/0/12 |

about the problem set used for analysis of algorithmic efficiency and accuracy, and section B comprises of tabular and graphical data to reinforce the claim of superiority of the proposed approach.

### 4.1 Problem Set Description

The set of objective functions considered for testing the proposed algorithm and compare its perfor-mance against classical DE and its variants PSODE and JADE have been taken from the CEC 2017 set of benchmark functions. Exhaustive comparisons and analysis have been depicted on dimensions D = 10, 30, 50 and 100 for a clear understanding of the strengths of the proposed algorithm. Objective functions $f_1 - f_3$ are simple unimodal functions and $f_4 - f_{10}$ are multimodal functions with a high number of local optima values. Functions $f_{11} - f_{20}$ are all hybrid functions using a combination of functions from $f_1 - f_{10}$. The set of composite function range from $f_{21} - f_{30}$ and merges the properties of the sub-functions better while incorporating the basic functions as well as hybrid functions to increase complexity while maintaining continuity around the global optima.

Table 4: Objective Function Value for Dimension: 100

| $f_{id}$ | DE | | JADE | | PSO-DE | | HIDE | |
|---|---|---|---|---|---|---|---|---|
| | best | mean | best | mean | best | mean | best | mean |
| $f_1$ | 3427212e+3 | 1380728e+4 | 141.263356 | 13516.69893 | 6067123.52 | 29751976.5 | **122.398748** | **11708.8236** |
| $f_2$ | 4.196e+84 | 1.547e+112 | 8.737e+74 | 2.543e+87 | **6.153e+66** | **3.211e+73** | 3.8835e+80 | 8.891e+114 |
| $f_3$ | 228808.969 | 262699.687 | 312244.360 | 332179.290 | 241427.723 | 257462.977 | **220765.083** | **251901.109** |
| $f_4$ | 1975.65115 | 2752.24606 | 539.386275 | 677.05465 | 777.314462 | 836.965399 | **531.169819** | **621.219143** |
| $f_5$ | 1223.53650 | 1286.15333 | 1249.19503 | 1307.11012 | 1248.41013 | 1310.88765 | **1068.11742** | **1272.47682** |
| $f_6$ | 651.65013 | 657.84974 | 654.70934 | 659.421427 | 656.87704 | 662.31841 | **642.33355** | **654.13275** |
| $f_7$ | 1614.00386 | 1920.79772 | 1367.06653 | 1536.35787 | **1311.84975** | **1534.20776** | 1562.37977 | 2076.70250 |
| $f_8$ | 1595.41873 | 1736.36737 | 1672.56784 | 1768.08243 | 1678.12726 | 1761.9405 | **1293.55211** | **1592.16298** |
| $f_9$ | 59726.5146 | 71986.0439 | 28906.9090 | 30336.7453 | 63640.3313 | 74961.2209 | **23466.5750** | **27067.0295** |
| $f_{10}$ | 12005.8897 | 14725.3483 | 14227.8019 | 15355.6218 | 12937.0278 | 14972.9507 | **11153.5868** | **13298.0921** |
| $f_{11}$ | 7540.6179 | 11481.2601 | 40447.5486 | 57228.6836 | **3521.90152** | **4544.80401** | 5380.43205 | 9916.34769 |
| $f_{12}$ | 529993877 | 1881773e+3 | 3893556.27 | 6415173.60 | 26105108.9 | 41876679.1 | **3680108.18** | **10059039.6** |
| $f_{13}$ | 7943.9249 | 508209.562 | 4622.69855 | 8892.77599 | 8246.51529 | 12675.8455 | **2976.84135** | 11376.9863 |
| $f_{14}$ | 728122.833 | 1329183.17 | **132194.795** | **365560.881** | 548410.338 | 941547.524 | 234045.940 | 867160.306 |
| $f_{15}$ | 2660.46578 | 181957.060 | **1799.50650** | 3362.50960 | 1899.07344 | 2914.44348 | 1976.78912 | 4485.4152 |
| $f_{16}$ | 4749.25466 | 5847.82673 | 4817.48373 | 5632.3022 | 3852.7000 | 5228.6635 | **3519.49494** | **4796.80272** |
| $f_{17}$ | 4397.49635 | 4958.41818 | 3842.20601 | **4450.17742** | 3790.72056 | 4730.99458 | **3582.78588** | 5463.21694 |
| $f_{18}$ | 1357845.39 | 1938893.27 | **146426.273** | **763318.822** | 1004224.20 | 2315010.2 | 631040.146 | 1335739.59 |
| $f_{19}$ | 2482.1701 | 26455.7069 | 2098.9496 | 4767.52953 | 2263.72515 | 3927.45994 | **2071.07706** | **3664.15987** |
| $f_{20}$ | 4968.49743 | 5436.60405 | 5231.02648 | 5690.74899 | 5109.46056 | 5781.30083 | **3627.77789** | **5228.43066** |
| $f_{21}$ | 3180.74665 | 3355.4783 | 2921.90012 | 3085.6922 | **2885.57408** | 3127.35683 | 2926.35039 | 3199.98618 |
| $f_{22}$ | 17808.8977 | 19562.9866 | 19213.3756 | 20278.9290 | 18695.5223 | 20167.41374 | **17548.3390** | **19547.1512** |
| $f_{23}$ | 4907.51964 | 5819.20786 | **3352.5569** | 4222.43689 | 3582.04355 | 4779.92124 | 3418.98320 | **3609.0985** |
| $f_{24}$ | 5173.24940 | 5946.12042 | 4060.95130 | 4095.42951 | **3801.36858** | **4042.42685** | 3998.05402 | 4216.82489 |
| $f_{25}$ | 4089.11891 | 4548.28576 | **3153.48541** | **3236.61784** | 3348.38226 | 3407.52658 | 3176.3038 | 3264.31853 |
| $f_{26}$ | 8557.49856 | 20159.1145 | 2900.07737 | 11924.79947 | 3021.13602 | 8682.03543 | **2900.00038** | **7867.5518** |
| $f_{27}$ | 3200.02335 | 3772.40915 | **3194.80921** | 3201.67073 | 3200.02417 | 3494.61813 | 3200.02354 | **3200.02395** |
| $f_{28}$ | 4947.74515 | 5948.21315 | **3295.12291** | **3340.28038** | 3456.82843 | 3542.57130 | 3300.80769 | 3354.71733 |
| $f_{29}$ | 6004.77442 | 7090.64254 | 5208.71172 | 5970.62868 | 5462.32863 | 6178.55906 | **4541.19547** | **5739.29154** |
| $f_{30}$ | 7798.10621 | 202435555 | **3584.97477** | 10674.2173 | 3920.32703 | **7139.46072** | 3850.31709 | 15318.5546 |
| w/t/l | 0/0/30 | 0/0/30 | 8/0/22 | 8/0/22 | 5/0/25 | 6/0/24 | 17/0/13 | 16/0/14 |

## 4.2 Parameter Settings

The work seeks to allow transparency in results by establishing a base for fair and clear comparisons in the analysis of the algorithms. The fixed values for the parameters have been depicted in table 1. The value of F and CR have been set as 0.5 and 0.9 for DE across all experiments, as recommended in the original document in (Storn and Price, 1995), (Mezura-Montes et al., 2006), (Brest et al., 2006). The parameters for JADE were selected as suggested in the initial work (Zhang and Sanderson, 2009). The values of parameters for PSO-DE have been retained from (Liu et al., 2010) as it is one of the more cited and prestiguous works. Also, we utilize the same parameter definitions for PSO as cited in this article by the initial authors in (Poli et al., 2007). The population size for initialised to 100 for all the algorithms as it is the uniformly recommended value by all of these papers. A total of 100 independant iterations were performed to obtain consistent result values to permit a uniform examination of the algorithm behaviour.

## 4.3 Numerical and Graphical Results

In tables 2-4, the best and mean values obtained for the population agents in the simulation runs have been reported, and the optimum values for each objective function have been highlighted in **bold**. For the sake of clarity, the comparison results in each table have been summarized in "w/t/l" format wherein w represents the number of objective functions where the algorithm outperforms all other algorithms, *t* specifies the number of objective functions where it is tied as the best algorithm for the objective function and l represents the number of test functions where it does not finish first. The utilization of the evaluation metric facilitates a definitive comparison of the different algorithms under consideration.

As represented in Table 2, On $D = 30$, HIDE achieved maximum number of wins in both best and mean case (17 and 18 respectively). JADE achieved second position with 8 and 9 wins in the best and mean case. The decent performance of JADE can be attributed to the adaptive nature of its parameter selection which enables enhancement of its convergence rate.

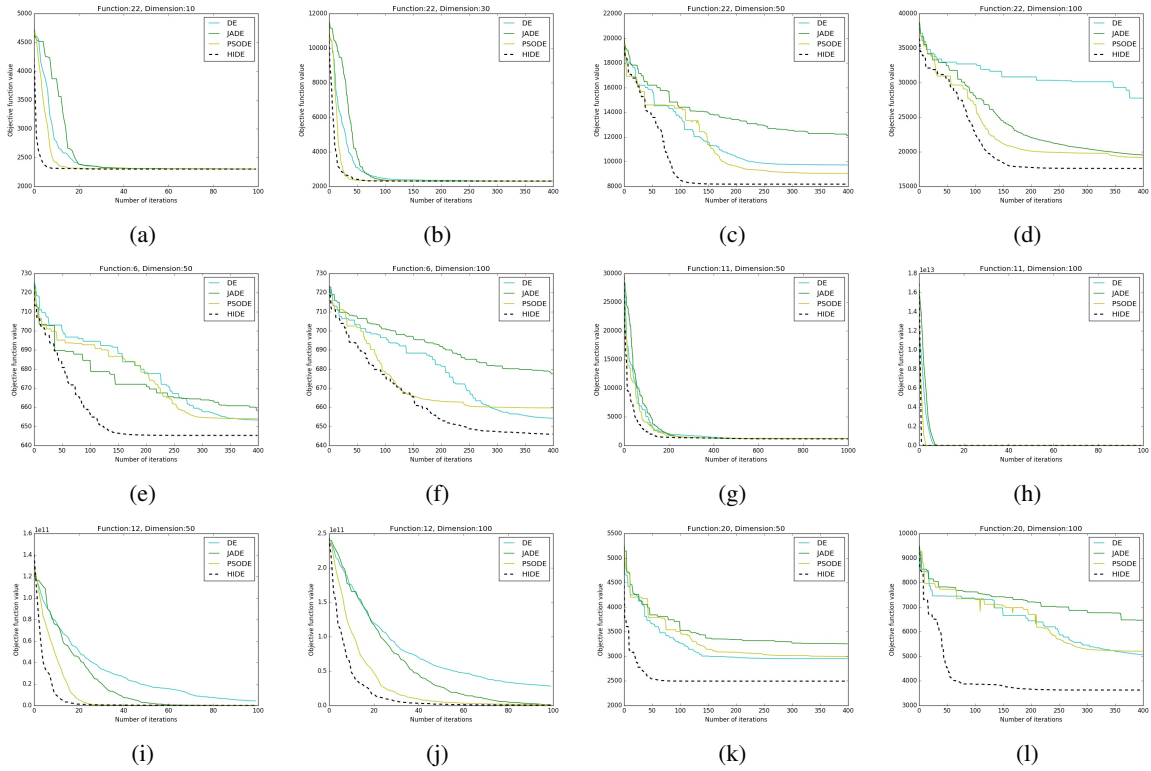

Figure 4: Comparative convergence profiles for test functions from CEC 2017 Benchmark over D = 10,30,50,100

The results for $D = 50$ and $D = 100$ (higher dimensions) have been summarized in tables 3 and 4. On $D = 50$, HIDE depicted exceptional performance, outperforming all other algorithms. It registered 17 wins in the best case and 18 wins in the mean case. Classical DE shows no wins in any case in high dimensional settings owing to its slow convergence rate and inability to attain global optimum thus highlighting the usefulness of the modifications introduced in the variants including HIDE. Similarly for $D = 100$, HIDE again outperforms all other algorithms by an appreciable margin. From a functional standpoint, It would be worthwhile to highlight that HIDE outperformed the other 3 compared algorithms on majority on the composite and hybrid functions, particularly on the higher dimensional settings. The efficiency of HIDE can be attributed to the hierarchical nature of crossover selection and concurrency in vector configurations at the higher hierarchy levels. The tabular results reinforce the fact that HIDE outperforms JADE, PSODE and DE. On close analysis, it can be witnessed that HIDE falls behind the other algorithms on a small fraction of unimodal functions such as $f_5, f_7$ on lower dimensions due to fast convergence during early stages of execution. However, the performance of higher dimensions, particularly on the more involved functions highlights utility for real world problems.

The tabular results are complemented through the graphical representations in Figure 4. For the sake of clarity, representations of higher dimensional problems span more number of iterations than those for lower dimensional settings. Analysis of the plots clearly depicts that HIDE shows better convergence rate as compared to other algorithms. As the analysis transcends to higher dimensional settings, the proposed approach outperforms the other algorithms on majority of the objective functions with respect to both convergence rate and optimality. the superiority of our algorithm in higher dimensions (50 and 100) is clearly evident from Figure 4 (c,d,g,h,k,l). Figure 4 (a,b,i,j) depict that for functions where HIDE and the other variants may depict similar trends on lower dimensions, HIDE eventually excels and surpasses them in higher dimensions in most scenarios. Almost all figures are representative of a faster convergence rate for HIDE on higher dimensions. This remarkable trait in HIDE enhances its utility for high dimensional problems where fast convergence to global optimum value is required, hence making it superior to the other considered algorithms and several variants of the DE algorithm.

## 5 Conclusion

Differential Evolution has been regarded as one of the most successful optimization algorithms and over the years, several variants have been proposed to enhance its convergence rate and performance. In the present work, we introduced a hierarchy influenced variant of the classical DE algorithm and modeled the same on the brain motor operation. The algorithm was characterized by global leader, local leaders and an effector population. The global leader and distributed local leaders interacted to facilitate gross motion via a greedy exploration strategy. The local leaders and their effectors interacted to control intricate motion for smooth convergence. A hierarchical crossover parameter was introduced to characterize the hierarchical transition between the two interactions. The influence of the vector configurations at the higher levels of hierarchy enabled the algorithm to avoid local minima in most objective functions. The same is complemented through our result observations wherein we significantly outperform several popular algorithm on complex multimodal functions in higher dimensional settings. Our proposed approach has sought to establish a viable tradeoff between fast optimization, robust convergence and low number of control parameters. The performance analysis of the algorithm highlights the particular effectiveness of the proposed approach on high dimensional hybrid and composite functions. The observed results provide sufficient motivation to extend the scope of the work to complex high dimensional real life problems including image enhancement, traveling salesman problem and flexible job-shop scheduling.